\pdfoutput=1

\documentclass[11pt]{article}

\usepackage[]{acl}

\usepackage{times}
\usepackage{arydshln} 
\usepackage{multirow}
\usepackage{booktabs}
\usepackage{latexsym}
\usepackage{graphicx}

\usepackage[T1]{fontenc}

\usepackage[utf8]{inputenc}

\usepackage{microtype}

\title{BioBART: Pretraining and Evaluation of \\ A Biomedical Generative Language Model}

\newcommand*\samethanks[1][\value{footnote}]{\footnotemark[#1]}

 \author{
Hongyi Yuan$^{1}$ \thanks{$\quad$Contributed equally.} \space\space
Zheng Yuan$^{1}$ \samethanks \space\space\space
Ruyi Gan$^{2}$ \space\space
Jiaxing Zhang$^{2}$ \space\space
Yutao Xie$^{2}$ \space\space
Sheng Yu$^{1}$ \thanks{$\quad$Corresponded author.} \\
$^{1}$Tsinghua University \space\space\space\space
$^{2}$International Digital Economy Academy\\
\texttt{\{yuanhy20,yuanz17\}@mails.tsinghua.edu.cn}\\
\texttt{\{ganruyi,zhangjiaxing,xieyutao\}@idea.edu.cn}\\
\texttt{syu@tsinghua.edu.cn}
}

\date{}

\begin{document}
\maketitle
\begin{abstract}
Pretrained language models have served as important backbones for natural language processing. Recently, in-domain pretraining has been shown to benefit various domain-specific downstream tasks. In the biomedical domain, natural language generation (NLG) tasks are of critical importance, while understudied. 
Approaching natural language understanding (NLU) tasks as NLG achieves satisfying performance in the general domain through constrained language generation or language prompting.
We emphasize the lack of in-domain generative language models and the unsystematic generative downstream benchmarks in the biomedical domain, hindering the development of the research community. 
In this work, we introduce the generative language model BioBART that adapts BART to the biomedical domain. 
We collate various biomedical language generation tasks including dialogue, summarization, entity linking, and named entity recognition.
BioBART pretrained on PubMed abstracts has enhanced performance compared to BART and set strong baselines on several tasks.
Furthermore, we conduct ablation studies on the pretraining tasks for BioBART and find that sentence permutation has negative effects on downstream tasks.
\end{abstract}

\section{Introduction}

Since the advent of ELMo \cite{elmo} and BERT \cite{Devlin2019BERTPO}, the new pretrain-then-finetune paradigm has brought great performance improvement and dominated the methodology research of the natural language processing (NLP) field. 
Previous research has illustrated that pretraining language models on the domain-specific corpora can improve the model performance on domain-specific tasks further \cite{Gururangan2020DontSP}. With the large-scale publicly accessible corpora from PubMed, researchers have already proposed biomedical domain pretrained language models such as BioBERT \cite{Lee2020BioBERTAP} and PubMedBERT \cite{pubmedbert} to aid the later research. 

Natural language generation (NLG) tasks such as dialogue system \cite{chao2017building} and question answering \cite{qasurvey} are of critical importance for the biomedical artificial intelligence research, and there is also a trend to approach natural language understanding as NLG tasks in the general domain \cite{sun2021paradigm,bartner}. For example, an entity retrieval task can be solved by constrained natural language generation \cite{GENRE}.
However, there exist two gaps in the research of the biomedical NLG. On the one hand, the architectures of the biomedical pretrained language models are almost all encoder-only transformers. Such architecture is incapable of generating natural languages auto-regressively. 
A decoder is necessary for language generation \cite{bertsumext}. 
On the other hand, there are very few in-domain generative language models for bio-medicine \cite{Phan2021SciFiveAT}. Models pretrained on biomedical corpora may further enhance the performance of current biomedical NLG methods.

To bridge the gaps mentioned above, we propose a biomedical auto-regressive generative language model, BioBART, pretrained on the biomedical corpora. 
In our work, we adopt BART (Bidirectional and Auto-Regressive Transformers), a generative pretrained language model which achieves state-of-the-art results on different NLG tasks in the general domain \cite{BART}. We continuously pretrain BART on PubMed abstracts to achieve biomedical domain adaption only using the text-infilling task. We also collate and evaluate BioBART on the existing biomedical NLG tasks. The in-domain BioBART outperforms BART model and sets strong baselines for several NLG tasks.

The main contributions of our work are summarized as follows\footnote{Our codes and pretrained checkpoints can be found at \url{https://github.com/GanjinZero/BioBART}.}:
\begin{enumerate}
    \item In aid of the research concerning the biomedical NLG tasks, we collate existing biomedical NLG tasks along with corresponding data and experimental settings. 
    The archived biomedical tasks will be released.
    
    \item We further analyze the influence of the pretraining task of sentence permutation in BART, and we find it brings degradation on the biomedical NLG tasks.
    
    \item 
    We evaluate our BioBART models on various NLG tasks and demonstrate the superb performance over BART.
    We will release the codes and weights to help reproduce our results.  
    
\end{enumerate}

\section{Related Work}

\subsection{Auto-regressive Language Model}

Most of the prestigious language models such as BERT, RoBERTa \cite{Liu2019RoBERTaAR} are auto-encoding transformers. The encoder-only architecture prevents the direct implementation of the seq2seq language generation. Several generative auto-regressive language models are proposed to mitigate the problem. The serial GPT models \cite{GPT1,GPT2,GPT3} adopt the decoder-only transformer architecture which is a left-to-right language model. They pretrain the models by auto-regressively predicting the upcoming word of sentences. UniLM1 \cite{UNILM1} and UniLM2 \cite{unilm2} implement attention masks to the transformer encoder to achieve unidirectional language modeling. They pretrain their models with a mixture of masked language modeling and auto-regressive language generation. T5 \cite{T5} and BART \cite{BART} apply the full transformer architecture, the encoder is used for input sequence encoding and the decoder is used for language generation. T5 and BART are both pretrained by denoising the corrupted corpora. Such models achieve many state-of-the-art results on various NLG tasks and some NLU tasks. 

\subsection{Biomedical Domain Pretraining}

Existing work has shown that pretraining the language models on the domain-specific corpora can bring better model transferability on the corresponding downstream tasks \cite{Gururangan2020DontSP}. There are endeavors to adapt language models to the specific domain. BioBERT \cite{Lee2020BioBERTAP} pretrained BERT model using biomedical corpora from PubMed abstracts and PubMed Central (PMC) full-text articles. BlueBERT \cite{Peng2020AnES} and clinicalBERT \cite{Huang2019ClinicalBERTMC} add electronic medical record (EMR) corpora from MIMIC-III \cite{Johnson2016MIMICIIIAF} to the pretraining data. Instead of continuous training from the general BERT checkpoint, SciBERT  \cite{Beltagy2019SciBERTAP} and PubMedBERT \cite{pubmedbert} are trained from scratch using scientific papers from Semantic Scholar \cite{semanticscholar} and PubMed articles respectively.  \cite{Shin2020BioMegatronLB} releases BioMegatron, a larger-size BERT-style language model pretrained on PubMed abstracts, PMC and MIMIC-III. The aforementioned work all use the model architecture of BERT. 
Other researchers are exploring different language models.

BioELMo \cite{Jin2019ProbingBE} is pretrained on biomedical corpora based on stacked bidirectional LSTM language model ELMo \cite{elmo}. BioELECTRA \cite{Kanakarajan2021BioELECTRAPretrainedBT} applies an adversarial training scheme consisting of a discriminator and a generator. They use PubMed abstracts and PMC articles as in-domain pretraining corpora.  BioMed-RoBERTa \cite{Gururangan2020DontSP} is initialized from RoBERTa \cite{Liu2019RoBERTaAR}, with additional training on the scientific papers from Semantic Scholar. Bio-lm \cite{biolm} is pretrained on data from PubMed, PMC, and MIMIC-III based on the RoBERTa model. 
KeBioLM \cite{kebiolm} uses Entity as Experts \cite{fevry-etal-2020-entities} model to inject biomedical entity knowledge into the language model, starting from the weights of PubMedBERT. Coder \cite{coder} and SapBERT \cite{sapbert} take advantage of the synonyms resource from biomedical knowledge base UMLS \cite{umls} and enhance the model with entity knowledge by contrastive pretraining. 

Due to the nature of model architecture, encoder-only language models have limited performance on the NLG tasks, such as summarization and question answering. In recent research, SciFive \cite{Phan2021SciFiveAT} is proposed for biomedical NLP tasks.
SciFive is pretrained on PubMed abstracts and PMC articles based on T5 architecture. 
While T5 is available for NLG tasks, SciFive is focused on evaluating NLU tasks.
Compared to SciFive, we choose to use BART as our model backbone and evaluate more on NLG tasks to leverage the power of decoders.


\subsection{Biomedical Natural Language Generation}

In the biomedical domain, most of the NLP tasks are natural language understanding (NLU) tasks. There are well-archived benchmarks for the evaluation of biomedical NLU, such as BLUE \cite{pubmedbert} and CBLUE \cite{zhang2021cblue}. NLG tasks are relatively less studied. \cite{ju2020CovidDialog} collects the patients and doctors' dialogues and forms a benchmark for Covid-19 related dialogue system. \cite{mediqa} is an annual biomedical NLP competition containing NLG tasks such as medical question (or answer) summarization and figure captions.

Moreover, with the success of GPT-3, there is a novel trend that unifies all the NLP tasks as NLG tasks \cite{nlpdeca,GPT3}. The traditional NLU tasks can be approached by constrained language generation. Much attention is paid on the NLG methods recently. In the biomedical domain, entities are of primary concern. GENRE \cite{GENRE}, \citet{oursnaaclshort} and BARTNER \cite{bartner} reach the new state-of-the-art by auto-regressive language model on entity linking and named entity recognition tasks. Such methods can be adapted to the biomedical domain. 


\section{Biomedical Domain Pretraining}

BART is a sequence-to-sequence model with a bi-directional encoder and a left-to-right auto-regressive decoder. The model architecture is consistent with the Transformers \cite{transformer} except for changing the ReLU activation functions to GeLUs \cite{gelu}. BART is pretrained by denoising the corrupted input documents. The work ablates five different types of corruption noise: text masking, text deletion, text infilling, sentence permutation, and document rotation. As a result, the pretraining documents are corrupted in two ways: 1) \textbf{Text Infilling}: For each document, a number of token spans are sampled, and each sample span is replaced with a single mask token. 2) \textbf{Sentence Permutation}: A document is split into sentences and sentences are shuffled in random orders. The pretraining objective is to minimize the negative log-likelihood of the original documents. 

Prior work has shown that continuous-pretrained models can get competitive results compared with those trained from scratch \cite{pubmedbert}. In our work, we continuously pretrain BART on the biomedical domain corpora. We revisit the methods to corrupt input texts. BART keeps the sentence permutation noise because of the significant performance gain on the summarization task, although this noise may lead to slight degradation on other tasks. We run further ablation studies on various biomedical NLG tasks. We show that the model pretrained without sentence permutation has better performance. Further details are listed in Section 5.5. Therefore we only implement the text infilling task to corrupt input texts for pretraining BioBART. 





\section{Generative Downstream Task}
In this section, we introduce the generative downstream tasks in the biomedical domain. We will conduct experiments on these tasks to illustrate the performance of the domain-specific BioBART. 

\subsection{Dialogue System}

A medical dialogue system aims to imitate the human doctor to communicate with human patients in a natural way. Based on the BART-style model, the patients' primitive descriptions and dialogue histories are used as inputs to the model, then the model auto-regressively generates the replies as outputs. The task is trained and evaluated in a sequence-to-sequence fashion. 

\subsection{Abstractive Summarization}

Summarization is a classical NLP task. It is important for healthcare to concisely summarize knowledge-rich biomedical
documents. Technically, there are abstractive and extractive approaches to generate better summaries. With the help of large pretrained language models, abstractive summarization methods outperform extractive methods in summary diversity and conciseness \cite{zhang2019pegasus,gsum}. The abstractive summarization is naturally an NLG task. We follow the BART \cite{BART} work and evaluate our BioBART on the biomedical summarization tasks in the same fashion. The input documents are encoded by the model encoder and the summaries are generated by the decoder auto-regressively. 

\subsection{Entity Linking}

Entity linking is a task that maps entity mentions in texts to its standard entity concepts. Traditional entity linking methods use language models to encode entity concepts from knowledge bases(e.g. UMLS) and mentions into the same dense space and disambiguate mentions by vector similarity. The large memory footprint requirements and difficult model training hinder the development of such methods. \citet{GENRE} proposes GENRE which uses generative language models to disambiguate entity mentions by auto-regressively generating the standard concept names conditioned on the inputs. \cite{oursnaaclshort} achieves state-of-the-art entity linking performance on various biomedical entity linking datasets by generative methods. We include this leading-edge method to show the superior performance of BioBART.

\subsection{Named Entity Recognition}
Named entity recognition (NER) is a critical task in the biomedical NLP community which extracts biomedical-related entities from texts.
Nested and discontinuous entities widely exist in biomedical papers and EMR due to the multi-granularity semantic meanings and complex syntax structures \cite{yuan2020unsupervised}.
Well-used sequential labelling framework in NER \cite{lample2016neural} is not directly fitted for nested and discontinuous NER \cite{finkel2009nested}.
\citet{bartner} propose BARTNER to model nested and discontinuous NER into seq2seq task by inputting sentences and outputting entities with their entity types one by one.
The generative approach of BARTNER achieves state-of-the-art performance on nested and discontinuous NER datasets, and we will use it to evaluate our proposed BioBART can further enhance the performance.

\section{Experiments}

\subsection{Pretraining}
\paragraph{Pretraining Corpora}

There are two main sources of biomedical corpora: PubMed abstracts, PMC articles. In the prior work \cite{pubmedbert}, training on both corpora surprisingly leads to a slight degradation in performance compared to solely training on PubMed abstracts. 
Therefore, we only use PubMed abstracts as the pretraining corpora. 
The corpora contain about 41 GB of biomedical research paper abstracts on PubMed.

\paragraph{Pretraining Setup}

We continuously pretrain both large and base versions of BART for 120k steps with a batch size of 2560. We use the same vocabulary as BART to tokenize the texts. Although the input length limitation of BART is 1024, the tokenized PubMed abstracts rarely exceed 512. Therefore, for the sake of training efficiency, we truncate all the input texts to 512 maximum length. We mask 30\% of the input tokens and the masked span length is determined by sampling from a Poisson distribution ($\lambda=3$) as used in BART. We use a learning rate scheduler of 0.02 warm-up ratio and linear decay. The learning rate is set to 1e-4. We train the base version of BioBART on 2 DGX with 16 40GB A100 GPUs for about 100 hours and the large version of BioBART on the same devices for 168 hours with the help of the open-resource framework DeepSpeed \cite{deepspeed}. 

\subsection{Dataset for Downstream Task}

\begin{table*}[ht]
\small 
\centering
\resizebox{2.1\columnwidth}{!}{
\begin{tabular}{cllllllllc}
\hline \hline
Task & Dataset & Train & Dev & Test & Dataset & Train & Dev & Test & Metric\\
\hline 
\multirow{2}{*}{Dialogue}&\multirow{2}{*}{CovidDialog}  &\multirow{2}{*}{490}&\multirow{2}{*}{63}&\multirow{2}{*}{61}&&&&&Rouge,BERTscore,\\
&&&&&&&&&BLEU\\
\hdashline
\specialrule{0em}{1pt}{1pt}
\multirow{3}{*}{Summarization}&MeQSum &500&-                                                                        &500&MEDIQA-ANS &38,166&174&552&\multirow{3}{*}{Rouge, BERTscore} \\
&iCliniq &24,851&3,105&3,108&MEDIQA-QS&1,000&50&100  \\
&HealthCareMagic &181,122&22,641&22,642& MEDIQA-MAS&1,104&50&80 \\
\specialrule{0em}{1pt}{1pt}
\hdashline
\specialrule{0em}{1pt}{1pt}
\multirow{3}{*}{Entity Linking}&MedMentions &122,241&40,884&40,157&NCBI&5,784&787&960&\multirow{3}{*}{Recall@1,@5}   \\
&BC5CDR &9,285&9,515&9,654&COMETA &13,489&2,176&4,350 \\
&AskAPatients &16,826&1,663&1,712  \\
\specialrule{0em}{1pt}{1pt}
\hdashline
\specialrule{0em}{1pt}{1pt}
\multirow{2}{*}{NER}&ShARe13 &5,146&669&5,333&ShARe14&10,380&771&7,922&\multirow{2}{*}{Entity-level F1 score}\\
&CADEC &4,430&898&990&GENIA&50,509&-&5,506\\

\hline \hline
\end{tabular}
}
\caption{The statistics of the datasets for biomedical generative tasks. The counts for NER are entity counts.}
\label{tab:dataset}
\small 
\end{table*}

\begin{table*}[ht]
\small 
\centering
\begin{tabular}{lccccc}
\hline 
 & \multicolumn{5}{c}{Covid19-Dialogue} \\
\cline{2-6}  
\textbf{Model} & Rouge-1 & Rouge-2 & Rouge-L & BLEU & BERTscore \\
\hline\specialrule{0em}{1pt}{1pt}
BART BASE &27.24 & 12.31 & 25.66 & 10.36 & 0.852 \\
BioBART BASE&28.14&\underline{12.77}&26.32&\underline{11.40}&0.849 \\
\specialrule{0em}{1pt}{1pt}\hdashline\specialrule{0em}{1pt}{1pt}
BART LARGE &\textbf{29.02}&12.08&\underline{26.93}&10.96&\textbf{0.852}\\
BioBART LARGE &\underline{28.81}&\textbf{13.79}&\textbf{26.96}&\textbf{12.05}&\underline{0.850}\\
\specialrule{0em}{1pt}{1pt}\hdashline\specialrule{0em}{1pt}{1pt}
State-of-the-art&-&-&-&7.60&-\\
Source&-&-&-&\cite{yang2020generation}&-\\
\hline
\end{tabular}
\caption{The main results on Dialogue System task.}
\label{tab:dialogue}
\small 
\end{table*}

\begin{table*}[ht]
\small 
\centering
\resizebox{2.1\columnwidth}{!}{
\begin{tabular}{lcccccccc}
\hline
 & \multicolumn{2}{c}{iCliniq} && \multicolumn{2}{c}{HealthCareMagic}  && \multicolumn{2}{c}{MEDIQA-QS}\\
\cline{2-3}\cline{5-6}\cline{8-9} 
\textbf{Model} & Rouge-1/2/L&BERTscore&& Rouge-1/2/L&BERTscore&& Rouge-1/2/L&BERTscore  \\
\hline
\specialrule{0em}{1pt}{1pt}
BART BASE &\underline{61.43}/\underline{48.68}/\textbf{59.71}&\textbf{0.941}&&46.81/\underline{26.19}/\underline{44.34}&\underline{0.918}&&28.82/10.99/26.99&0.896\\
BioBART BASE &61.07/48.47/\underline{59.42}&\textbf{0.941}&&46.67/26.03/44.11&\underline{0.918}&&30.12/11.28/27.44&0.898\\
\specialrule{0em}{1pt}{1pt}\hdashline\specialrule{0em}{1pt}{1pt}
BART LARGE &59.87/47.01/58.12&0.938&&\textbf{47.24}/\textbf{26.54}/\textbf{44.68}&\textbf{0.919}&&29.97/10.64/28.41&\underline{0.901} \\
BioBART LARGE &60.32/47.98/58.69&\underline{0.940}&&46.54/26.14/44.23&\textbf{0.919}&&\underline{31.97}/\underline{12.39}/\underline{29.70}&\textbf{0.903}  \\
\specialrule{0em}{1pt}{1pt}\hdashline\specialrule{0em}{1pt}{1pt}
State-of-the-art& \textbf{62.3}/\textbf{48.7}/58.5 & -&&\underline{46.9}/24.8/43.2 &-&&\textbf{35.14}/\textbf{16.08}/\textbf{31.31}&-\\
Source&\multicolumn{2}{l}{\cite{mrini-etal-2021-gradually}}&&\multicolumn{2}{l}{\cite{mrini-etal-2021-gradually}}&&\multicolumn{2}{l}{\cite{mediqa}}\\
\specialrule{0em}{3pt}{3pt}
 & \multicolumn{2}{c}{MEDIQA-MAS} && \multicolumn{2}{c}{MEDIQA-ANS(Pages)}&& \multicolumn{2}{c}{MeQSum}\\
\cline{2-3}\cline{5-6}\cline{8-9}
\textbf{Model} & Rouge-1/2/L&BERTscore&& Rouge-1/2/L&BERTscore && Rouge-1/2/L&BERTscore \\
\hline
\specialrule{0em}{1pt}{1pt}
BART BASE & \underline{31.63}/9.98/\underline{27.85}&\underline{0.859}&&19.10/6.77/16.90&0.851&&52.93/35.79/50.46&0.927\\
BioBART BASE &\textbf{32.90}/\underline{11.28}/\textbf{29.26}&\textbf{0.861}&&18.97/7.46/16.77&0.850 && 53.75/36.50/\underline{51.27}&\underline{0.929}\\
\specialrule{0em}{1pt}{1pt}\hdashline\specialrule{0em}{1pt}{1pt}
BART LARGE &29.32/9.00/26.14&0.857&&21.52/\underline{9.31}/\underline{19.15}&\underline{0.853}&&53.68/36.80/51.05&0.928 \\
BioBART LARGE &{30.60}/10.37/27.04&\textbf{0.861}&&\underline{21.58}/\textbf{9.34}/\textbf{19.18}&\textbf{0.857}&& \textbf{55.61}/\textbf{38.11}/\textbf{53.15}&\textbf{0.933} \\
\specialrule{0em}{1pt}{1pt}\hdashline\specialrule{0em}{1pt}{1pt}
State-of-the-art&32.15/\textbf{16.21}/19.10&-&&\textbf{23.07}/ 5.41/15.35&-&&\underline{54.5}/\underline{37.9}/50.2&-\\
Source&\multicolumn{2}{l}{\cite{mediqa}}&&\multicolumn{2}{l}{\cite{laskar2021domain}}&&\multicolumn{2}{l}{\cite{mrini-etal-2021-gradually}}\\
\hline
\end{tabular}
}
\caption{The main results on Summarization tasks.}
\label{tab:sum2}
\small 
\end{table*}

\begin{table*}[h]
\small 
\centering
\begin{tabular}{lccccc}
\hline 
 & MedMentions & BC5CDR & NCBI & COMETA & AAP \\
\textbf{Model} & Recall@1/@5 & Recall@1/@5 & Recall@1/@5 & Recall@1/@5 & Recall@1/@5 \\
\hline
\specialrule{0em}{1pt}{1pt}
BART BASE &69.77/84.59 &91.56/94.89 &88.54/95.31 &78.34/87.40 &86.37/94.29  \\
BioBART BASE &71.15/\textbf{86.22}&\underline{93.01}/\underline{95.59}&89.27/95.31&79.63/88.64&87.51/94.92\\
\specialrule{0em}{1pt}{1pt}\hdashline\specialrule{0em}{1pt}{1pt}
BART LARGE &71.49/84.95 & 92.48/95.26 & \underline{90.21}/\underline{95.52}& \underline{80.70}/\underline{88.65} &88.79/\textbf{96.59} \\
BioBART LARGE &\underline{71.78}/\underline{85.42} &\textbf{93.26}/\textbf{95.74}&89.90/\textbf{95.63}&\textbf{81.77}/\textbf{88.87}&\textbf{89.40}/\underline{95.76}\\
\specialrule{0em}{1pt}{1pt}\hdashline\specialrule{0em}{1pt}{1pt}
State-of-the-art&\textbf{74.6}/ -&91.9/ -&\textbf{92.4}/ -&80.1/ -&\underline{89.0}/ -\\
Source&\cite{dataintegration}&\cite{dataintegration}&\cite{rescnn}&\cite{rescnn}&\cite{sapbert}\\
\hline
\end{tabular}
\caption{The main results on Entity Linking tasks.}
\label{tab:main:linking}
\small 
\end{table*}

\begin{table}[h]
\small 
\centering
\resizebox{1\columnwidth}{!}{
\begin{tabular}{lccccc}
\hline 
& ShARe13 & ShARe14 & CADEC & GENIA \\
\textbf{Model} & F1 & F1 & F1 & F1 \\
\hline
\specialrule{0em}{1pt}{1pt}
BART BASE  & 76.63 & 77.87 & 68.37 & 78.06 \\
BioBART BASE & 78.78 & 79.17 & 68.39 & 78.43\\
\specialrule{0em}{1pt}{1pt}\hdashline\specialrule{0em}{1pt}{1pt}
BART LARGE & 79.69 & 80.34 & \underline{70.64} & 78.93 \\
BioBART LARGE &\underline{80.75}&\underline{80.41}&70.53&\underline{79.93} \\
\specialrule{0em}{1pt}{1pt}\hdashline\specialrule{0em}{1pt}{1pt}
State-of-the-art & \textbf{82.52} & \textbf{81.75} & \textbf{73.21} & \textbf{81.39}\\
Source&\multicolumn{4}{c}{\cite{w2ner}}\\
\hline
\end{tabular}
}
\caption{The main result on NER tasks.}
\label{tab:ner}
\small 
\end{table}

\subsubsection{Dialogue System}
\paragraph{CovidDialog} \cite{ju2020CovidDialog} Concerning the widespread Coronavirus disease 2019 (COVID-19) pandemic, the CovidDialog dataset is proposed to facilitate the development of dialogue system providing COVID-related consultations to people. The dataset is collected from online healthcare forums.
It contains 603 consultations about COVID-19 and other related pneumonia, having 1232 utterances in total. Each consultation starts with a description related to patients' medical conditions, then followed the conversation between a doctor and a patient. 

\subsubsection{Abstractive Summarization}
\paragraph{iCliniq, HealthCareMagic} Both datasets are extracted from MedDialog \cite{zeng-etal-2020-meddialog} dataset, collected from the online healthcare platform. iCliniq contains 31,062 samples and HealthCareMagic contains 226,405 samples. Each sample is comprised of a summary and corresponding dialogues between a patient and a doctor. HealthCareMagic’s summaries are more abstractive and are written in a formal style, unlike iCliniq’s patient-written summaries.
We follow the previous work \cite{mrini-etal-2021-gradually} for training, developing, and testing data separations of both datasets.


\paragraph{MeQSum} \cite{MeQSum} The dataset is created for better medical question summarization because the original patients' questions are verbose, causing difficulty for the question-answering system. The dataset contains 1000 patients' health questions selected from a collection distributed by the U.S. National Library of Medicine \cite{Kilicoglu2018SemanticAO}. Each question is annotated with a question summarization by medical experts.

\paragraph{MEDIQA-ANS} \cite{mediqaans} When feeling discomfort, people may turn to the internet for the answers to their medical questions. The raw searching result may be obscure for even medical experts. The dataset is proposed to emphasize the need for a medical answer summarization system in aid of better understanding biomedical materials. It consists of 156 health questions, corresponding answers to these questions, and expert-created summaries (both abstractive and extractive) of these answers. Following the paper, we use BioASQ \cite{bioasq} to construct training data, MedInfo \cite{medinfo} for validation, and the whole MEDIQA-ANS dataset for testing.

\paragraph{MEDIQA-QS, MEDIQA-MAS} Both datasets are derived from the MEDIQA 2021 Tasks \cite{mediqa}. MEDIQA-QS dataset aims to incentivize the development of new summarization approaches that address specifically the challenges of long and complex health questions. The dataset provides the validation and test sets, and MeQSum dataset is used as the training set. MEDIQA-MAS aims to prompt research that simultaneously aggregates and summarize the different relevant answers to a medical question. This dataset provides the validation and test sets, and MEDIQA-ANS dataset comprises the training set.

\subsubsection{Entity Linking}
\paragraph{MedMentions} \cite{medmentions} MedMentions is a large-scale biomedical entity recognition dataset. The commonly used St21pv subset contains 4,392 PubMed abstracts, and over 350,000 mentions are linked to concepts of 21 selected semantic types in UMLS \cite{umls}.

\paragraph{BC5CDR} \cite{li2016biocreative} BC5CDR is a benchmark for biomedical entity linking. 1500 PubMed article abstracts are annotated with 4409 chemicals, 5818 diseases entities, and 3116 chemical-disease interactions. MeSH ontology, a subset of UMLS is used to annotate entities. We follow most recent work \cite{clustering,dataintegration} for data pre-processing.

\paragraph{NCBI} \cite{dougan2014ncbi} The dataset is built from 793 PubMed abstracts. It consists of 6892 annotated disease mentions of 790 unique disease concepts. The annotators label all the mentions to concepts in MEDIC ontology \cite{davis2012medic}. MEDIC is a medical dictionary that merges the diseases concepts, synonyms, and definitions in MeSH and OMIM and is composed of 9700 unique diseases. We follow BioSyn \cite{sung2020biomedical} to process data and construct dataset splits. 

\paragraph{COMETA} \cite{cometa} COMETA is derived from the online publicly available and anonymous health discussion on Reddit. It consists of 20k English biomedical entity mentions expert-annotated with concepts from SNOMED CT. We use the “stratified (general)” split and follow the training and evaluation procedures of SapBert \cite{sapbert} and ResCNN \cite{rescnn}.

\paragraph{AskAPatient} \cite{limsopatham-collier-2016-normalising} It contains 8,662 phrases from social media. Each phrase can be mapped to one of the 1,036 medical concepts from SNOMED-CT and AMT (the Australian Medicines Terminology). The samples in AskAPatient do not include contextual information. We follow \citet{sung2020biomedical} and \citet{limsopatham-collier-2016-normalising} for data pre-processing and apply the 10-fold evaluation protocol.

\subsubsection{Named Entity Recognition}
\paragraph{ShARe13, ShARe14, CADEC}  These three datasets annotate discontinuous adverse drug events entities. The main difference is the annotated data of ShARe tasks \cite{pradhan2013task,mowery2014task} comes from MIMIC-II, and CADEC \cite{karimi2015cadec} comes from social media. There is only one entity type for these datasets. We follow \citet{bartner} for dataset preprocess.

\paragraph{GENIA} \cite{kim2003genia} GENIA annotates 2000 MEDLINE abstracts with biological entities.
Entities can be nested with others. We follow \cite{lin2019sequence} to combine fine-grained entity types into 5 coarse-grained entity types and to construct dataset splits. 

All the aforementioned datasets are in English. The statistical overview of the aforementioned datasets is listed in Table \ref{tab:dataset}.

\subsection{Fine-tuning details}

\paragraph{Dialogue}
We use BioBART as the dialogue system model. The dialogue history is fed into the encoder and the decoder generates the response auto-regressively. We apply the negative log-likelihood function as the training objective with respect to the reference dialogue response. We fine-tune the model with learning rate 5e-5 for the base version and 1e-5 for the large version for 20 epochs. We run evaluations on the validation set at the end of each epoch and use the checkpoint with the best validation performance for testing. During inference, we use beam search of size 5 to sample responses from the model's outputs. We use Rouge-1/2/L \cite{lin-2004-rouge}, BLEU \cite{bleu} and BERTscore \cite{zhang2020bertscore} as our evaluation metrics. RoBERTa-large \cite{Liu2019RoBERTaAR} is used as scorer in BERTscore.

\paragraph{Summarization}
Similarly, for summarization, the encoder takes the documents as input, and the decoder generates the corresponding summarizations. We minimize the log-likelihood objective to fine-tune the model and apply beam search for inference. Across different summarization datasets, the beam size is set to 5 and we use no length penalty. We fine-tune the model with learning rate 5e-5 for the base version and 1e-5 for the large version for 6 epochs. We run evaluations on the validation set at the end of each epoch and use the checkpoint with the best validation performance for testing. We apply the commonly used Rouge-1/2/L and BERTscore for evaluation metrics. The large version of RoBERTa is used as the scorer in BERTscore.

\paragraph{Entity Linking} We follow the method and experimental settings in \citet{oursnaaclshort} to implement the generative model for biomedical entity linking tasks. We do not apply knowledge-base guided pre-training proposed in \citet{oursnaaclshort}. The documents with the positions of mentions marked are fed into the encoder and the decoder outputs the corresponding synonyms in the knowledge base directly. We use the top1 and top5 recall (Recall@1 and Recall@5) as the evaluation metrics.

\paragraph{NER} We use BARTNER \cite{bartner} as our model. The target type for BARTNER is \textit{word} (i.e. output first BPE of each word in entities). We use the parameters selected by \citet{bartner} for all pretrained models and fine-tune for 30 epochs. Entity-level F1 is used as the metric.
\begin{table*}[t]
\centering
\resizebox{2.1\columnwidth}{!}{
\begin{tabular}{lcccccccc}
\hline
 & \multicolumn{2}{c}{CovidDialogue} && \multicolumn{2}{c}{MeQSum} && \multicolumn{2}{c}{MEDIQA-MAS}  \\
\cline{2-3} \cline{5-6} \cline{8-9}
 & Rouge-2/L & BLEU && Rouge-2/L & BERTscore &&Rouge-2/L & BERTscore \\
 \specialrule{0em}{1pt}{1pt}\hdashline\specialrule{0em}{1pt}{1pt}
BART BASE & \textbf{12.31}/25.66 & 10.36 && 35.79/50.46&0.927&&9.98/\textbf{27.85} &0.859 \\
\specialrule{0em}{1pt}{1pt}\hdashline\specialrule{0em}{1pt}{1pt}
\ \ w/ TI \& SP&10.90/25.46&10.46&&34.93/49.28&0.926&&9.04/27.43&0.859 \\
\ \ w/ TI & 11.81/\textbf{25.79}&\textbf{12.79}&&\textbf{37.14}/\textbf{51.71}&\textbf{0.929}&&\textbf{10.66}/27.65&\textbf{0.862}\\
\specialrule{0em}{1pt}{1pt}\hdashline\specialrule{0em}{1pt}{1pt}
 &\multicolumn{2}{c}{MedMentions} && \multicolumn{2}{c}{COMETA} && ShARe13 & CADEC \\
\cline{2-3} \cline{5-6}
 & Recall@1&Recall@5 && Recall@1&Recall@5 && F1 & F1 \\
 \specialrule{0em}{1pt}{1pt}\hdashline\specialrule{0em}{1pt}{1pt}
BART BASE &69.77&84.59 &&78.34&87.40 && 76.63 & 68.37 \\
\specialrule{0em}{1pt}{1pt}\hdashline\specialrule{0em}{1pt}{1pt}
\ \ w/ TI \& SP&70.76&\textbf{85.69}&&79.31&87.93&&77.01&67.98 \\
\ \ w/ TI &\textbf{70.82}&85.68&&\textbf{79.43}&\textbf{88.28}&& \textbf{78.03}&\textbf{69.22}\\

\hline
\end{tabular}
}
\caption{Ablation study for pretraining tasks. TI is short for Text Infilling and SP is short for Sentence Permutation.}
\label{tab:ablation}
\end{table*}
\subsection{Main Result}
\begin{table*}[t]
\scriptsize
\centering
\resizebox{2.1\columnwidth}{!}{
\begin{tabular}{p{0.6\columnwidth}|p{0.6\columnwidth}|p{0.6\columnwidth}}
\textbf{Context(Document)} & \textbf{BART Response(Summary)} & \textbf{BioBART Response(Summary)}\\
\specialrule{0em}{1pt}{1pt}\hline
Description: I have type 1 diabetes that I try hard to control, but i struggle with. Are my chances high for poor outcomes with the virus? I’m in such a panic about it.&In brief: diabetes. \textcolor{red}{Your risk of complications with covid-19 is very low.} If you are well controlled your risk should not be elevated. If your diabetes is uncontrolled you should not have a particularly high risk.&Diabetes and Covid-19. If you are a type 1 diabetic, \textcolor{blue}{you are at high risk for poor outcomes with the virus.} Your risk is very high. You should be aware of your risk factors and monitor your glycemic control closely. \\
\specialrule{0em}{1pt}{1pt}\hline\specialrule{0em}{1pt}{1pt}
SUBJECT: treatment for \textcolor{blue}{dry mouth caused by necessary medicine} MESSAGE: My provider can't help (I asked.) \textcolor{blue}{I am intolerant of all the "sugar alcohols"} such as maltilol, ...&What are the treatments for \textcolor{red}{dry mouth caused by sugar alcohols}?&What are the treatments for dry mouth?\\
\specialrule{0em}{1pt}{1pt}\hline\specialrule{0em}{1pt}{1pt}
Are there any new treatments for orthostatic tremor? ... \textcolor{red}{Deep brain stimulation is sometimes done when drugs cannot control a severe, disabling tremor. Sometimes essential tremors or tremors due to Parkinson disease or another disorder require such treatment.} ... \textcolor{blue}{There is no cure for most forms of tremor, but there are treatments to help manage symptoms.} ...
&\textcolor{blue}{There is no cure for tremor, but there are treatments to help manage the symptoms.} &\textcolor{blue}{There is no cure for tremor, but there are treatments to help manage the symptoms.} The most common ... help relieve symptoms. \textcolor{red}{Deep brain stimulation is sometimes done when drugs cannot control a severe, disabling tremor. Sometimes essential tremors or tremors due to Parkinson disease or another disorder require such treatment.}\\
\specialrule{0em}{1pt}{1pt}\hline
\end{tabular}
}
\caption{Example dialogue and summaries from the fine-tuned BART and BioBART model
excepts of showing the references. The key information and differences are highlighted with colors.}
\label{tab:case}
\end{table*}
In this section, we present the base and large version of BioBART on various generation tasks. We compare our in-domain BioBART with BART to illustrate the effectiveness of domain adaption. We also compare with the existing state-of-the-art results on each dataset to shed light on the superior performance of BioBART. The experimental results are shown in Table \ref{tab:dialogue}-\ref{tab:ner}. The best and the second-best scores are highlighted with bold numbers and underlines respectively.

\paragraph{Dialogue} We evaluate biomedical dialogue response generation on CovidDialog. For both base and large version, BioBART shows improvement on the automatic metric Rouge. The large BioBART outperforms BART by 1.71 on Rouge-2 and 0.03 on Rouge-L . Our evaluations surpasses the current state-of-the-art on BLEU score by 4.45.

\paragraph{Summarization} We present broad experimental results on biomedical summarization datasets. From Table \ref{tab:sum2}, BioBART has competitive or even superior performance on the task. Except for iCliniq and HealthCareMagic, we see consistent improvement on different datasets for both sizes of BioBART. For MeQSum, BioBART large exceeds BART large for 1.93/1.31/2.1 on Rouge-1/2/L and even outperforms the current state-of-the-art. The possible reason that biomedical in-domain pretraining fails on iCliniq and HealthCareMagic is that both datasets are built upon a clinical corpus. There still exists a domain-shifting problem for BioBART pretrained on biomedical scientific articles from PubMed.

On dialogue and summarization tasks, there are minor changes in BERTscore for different models. This is possible because the metric is calculated by other pretranined language models. The implemented RoBERTa may suffer from biomedical domain-shifting and cannot quantify the model performance accurately. 

\paragraph{Entity Linking} The results on biomedical entity linking tasks are shown in Table \ref{tab:main:linking}. For all the tasks, models finetuned based on BioBART have better performance. On AAP, BC5CDR, and COMETA, our results outperform the current discriminative state-of-the-art methods by 0.4, 1.67, and 1.36 points of Recall@1 respectively. 

\paragraph{NER} The performance improvement of BioBART on ShARe13, ShARe14, and GENIA is significant, while the increase on CADEC is mediocre. For the large models, BioBART improves entity-level F1 scores for 1.06 and 1 on ShARe13 and GENIA datasets. There are promising results for generative biomedical NER methods, while the gap with the current state-of-the-art NER method \cite{w2ner} is still salient.

\subsection{Ablation Study on Pretraining Task}

In this section, we test on pretraining with or without the sentence permutation task. We pretrain BART base following the same pretraining settings except for reducing the training step to 40k for efficiency. We fine-tuned the pretrained models on the downstream tasks. The ablation results are shown in Table \ref{tab:ablation}.

From the result, it is illustrated that the model pretrained on isolated text infilling task performs the best. The sentence permutation task downgrades the model's performance even for generative summarization and dialogue system tasks. 

\subsection{Generated example}

Here we demonstrate BioBART's performance qualitatively. In Table \ref{tab:case}, we present three generative examples on CovidDialog, MeQSum, and MEDIQA-ANS respectively. In the first example, we can see that BART generates an erroneous instruction of the influence of diabetes. BioBART injected with domain knowledge can correctly give the response. In the second, BART misunderstands the document where sugar alcohol is not the cause of dry mouth. BioBART generates an accurate and concise summary. In the final example, the MEDIQA-ANS document is rather long and BART fails to extract complete information (colored in red).
From the examples, we can conclude that BioBART has improvements on biomedical common sense and documents understanding.

\section{Conclusions}

In this work, we pretrain the biomedical domain generative language model BioBART. We also collect various publicly available benchmarks for biomedical generative tasks to prompt future research. Our experimental results show that continuous pretraining on PubMed abstracts helps the model with domain adaption. BioBART shows great improvements on different benchmarks and achieves competitive or superior results over the current state-of-the-art methods. We also release our pretraining and fine-tuning codes to facilitate future research for reproducibility. 

We will explore pretraining generative language models 1) on in-domain vocabularies and from scratch, 2) and with clinical corpora such as EMRs in MIMIC-III \cite{Johnson2016MIMICIIIAF} or PMC-Patients \cite{zhao2022pmcpatients} in the future studies. 

\section*{Acknowledgements}
We appreciate three anonymous reviewers for helpful comments. This work was supported by the National Natural Science Foundation of China (Grant No. 12171270), and the Natural Science Foundation of Beijing Municipality (Grant No. Z190024).

\bibliographystyle{acl_natbib}
\bibliography{acl}

\end{document}